\documentclass[conference]{IEEEtran}
\IEEEoverridecommandlockouts
\usepackage{cite}
\usepackage{amsmath,amssymb,amsfonts}
\usepackage{algorithmic}
\usepackage{graphicx}
\usepackage{textcomp}
\usepackage[dvipsnames]{xcolor}
\def\BibTeX{{\rm B\kern-.05em{\sc i\kern-.025em b}\kern-.08em
    T\kern-.1667em\lower.7ex\hbox{E}\kern-.125emX}}

\usepackage[
  bookmarks=true,
  pdfpagelabels=true,
  pdftitle={Paper Submission},
  hyperfootnotes=true,
  hyperindex=true,
  pageanchor=true,
  colorlinks=false,
  hidelinks
]{hyperref}

\usepackage{todonotes}
\usepackage[acronym]{glossaries}
\glsdisablehyper
\usepackage{threeparttable}
\usepackage{pifont}% http://ctan.org/pkg/pifont
\usepackage{multicol,multirow}
\usepackage[capitalise]{cleveref}
\usepackage{url}
\usepackage{soul}

\definecolor{ForestGreen}{RGB}{34,139,34}

\newcommand{\blue}[1]{{\color{black}#1}}

\newcommand{\oursTable}[1]{{\color{RoyalBlue}\textit{#1}}}

\usepackage{enumitem}
\usepackage{makecell}
\usepackage[  ]{siunitx}

\newacronym{pe}{PE}{Printed Electronics}
\newacronym{ml}{ML}{Machine Learning}
\newacronym{svm}{SVM}{Support Vector Machine}
\newacronym{sv}{SV}{Support Vector}
\newacronym{dt}{DT}{Decision Tree}
\newacronym{mlp}{MLP}{Multi-Layer Perceptron}
\newacronym{mac}{MAC}{Multiply-Accumulate}
\newacronym{nre}{NRE}{non-recurring engineering}
\newacronym{fmcg}{FMCG}{fast-moving consumer goods}
\newacronym{fsm}{FSM}{Finite State Machine}
\newacronym{mux}{MUX}{Multiplexer}
\newacronym{lut}{LUT}{Look-Up Table}
\newacronym{rom}{ROM}{Read-Only Memory}
\newacronym{bnn}{BNN}{Binary Neural Network}

\makeatletter
\newcommand*\titleheader[1]{\gdef\@titleheader{#1}}
\AtBeginDocument{%                                                               
  \let\st@red@title\@title
  \def\@title{%                                                                 
    \bgroup\normalfont\normalsize\centering\@titleheader\par\egroup
    \vskip1ex\st@red@title}
}
\makeatother

\title{\vspace{-8pt}Compact Yet Highly Accurate Printed Classifiers Using Sequential Support Vector Machine Circuits\vspace{-1ex}}
\author{\IEEEauthorblockN{
Ilias~Sertaridis, %\IEEEauthorrefmark{1},
Spyridon~Besias, %\IEEEauthorrefmark{1},
Florentia~Afentaki,
Konstantinos~Balaskas
and~Georgios~Zervakis
}
\IEEEauthorblockA{
Department of Computer Engineering \& Informatics, University of Patras, Patras, Greece
}
\IEEEauthorblockA{
\{st1072480, st1072524, afentaki, kompalas, zervakis\}@ceid.upatras.gr
\vspace{-3ex}
}}
\titleheader{\vspace{-20pt}To appear at the 2025 IEEE International Symposium on Circuits and Systems (ISCAS), May 25-28 2025, London, UK.}

\begin{document}

\maketitle

\begin{abstract}
Printed Electronics (PE) technology has emerged as a promising alternative to silicon-based computing.
It offers attractive properties such as on-demand ultra-low-cost fabrication, mechanical flexibility, and conformality.
However, PE are governed by large feature sizes, prohibiting the realization of complex printed Machine Learning (ML) classifiers.
Leveraging PE's ultra-low non-recurring engineering and fabrication costs, designers can fully customize hardware to a specific ML model and dataset, significantly reducing circuit complexity.
Despite significant advancements, state-of-the-art solutions achieve area efficiency at the expense of considerable accuracy loss.
Our work mitigates this by designing area- and power-efficient printed ML classifiers with little to no accuracy degradation.
Specifically, we introduce the first sequential Support Vector Machine (SVM) classifiers, exploiting the hardware efficiency of bespoke control and storage units and a single Multiply-Accumulate compute engine.
Our SVMs yield on average 6x lower area and 4.6\% higher accuracy compared to the printed state of the art.
\end{abstract}

\begin{IEEEkeywords}
Machine Learning, Support Vector Machines, Printed Electronics
\end{IEEEkeywords}

\bstctlcite{IEEEexample:BSTcontrol} % More than 6 authors --> use et. al

\section{Introduction}
\label{sec:introduction}
%Printed Electronics (\gls{pe}) have emerged as a transformative technology designed to address the limitations of traditional silicon-based systems, particularly in cost-sensitive and flexible applications~\cite{Bleier:ISCA:2020:printedml}.
\gls{pe} have emerged as a complement to silicon-based technology, meeting demands that are untouchable by the latter, such as mechanical flexibility, non-toxicity, conformability, and ultra-low cost~\cite{Bleier:ISCA:2020:printedml}. 
Target applications include smart packaging~\cite{smartpackaging2022, disposable:JSNB:2023}, forensics~\cite{salivary:Talanta:2020}, and accessible healthcare products and wearables~\cite{bodytemperature:sna:2020,pressuresensor:research:2022,wearable:adma:2022,Wearable:acssensors:2019,healthcare:Nanoscale:2024}.
Unlike conventional silicon technologies, which are constrained by high manufacturing and assembly costs, \gls{pe} leverage additive, mask-less printing processes that enable on-demand production of flexible circuits at significantly lower costs~\cite{cui2016printed}.
%\gls{pe} systems, although manufactured with micrometer-scale features and limited in performance compared to silicon-based devices, excel in domains requiring conformality, low power consumption, and non-toxicity~\cite{Armeniakos:TCAD2023:cross}.
%Importantly, recent advancements in \gls{pe} have enabled the development of low-voltage circuits powered by printed batteries, or even self-sustaining energy sources, further enhancing their potential in applications where traditional electronics fail to meet cost, stretchability, or flexibility demands~\cite{Mubarik:MICRO:2020:printedml}.

% Challenges for ML in PE - Bespoke ML classifiers
Such application domains stand to greatly benefit from the infiltration of \gls{ml} algorithms~\cite{Mubarik:MICRO:2020:printedml}.
%The ability to embed intelligence into low-cost printed devices could enable real-time monitoring, product tracking, and enhanced interactivity, all while keeping application costs low~\cite{Mubarik:MICRO:2020:printedml}.
However, \gls{pe} face significant challenges in implementing complex circuits such as \gls{ml} classifiers, due to the large feature sizes and the corresponding power and area overheads, thus hindering the ubiquitous integration of \gls{ml} in \gls{pe} applications.
To overcome these limitations, leveraging the low \gls{nre} and fabrication costs in PE, bespoke circuits have been proposed as a promising solution~\cite{Mubarik:MICRO:2020:printedml}.
They refer to fully customized designs tailored to a specific \gls{ml} model and dataset, optimizing hardware for a particular application.
Such tailored designs allow for significant reductions in power and area, by hardwiring the parameters of the \gls{ml} model into the circuit implementation.
This level of customization is infeasible in conventional lithography-based silicon technologies, due to elevated costs (e.g., \gls{nre}, maskset, etc.), particularly at low to moderate volumes;
even FPGA-based systems are constrained by the architecture, multiplexing, and routing of the underlying fabric.
%While traditional design techniques struggle to meet the constraints of \gls{pe}, bespoke architectures leverage the low \gls{nre} costs and high degree of customization, making it feasible to implement on-demand, resource-efficient \gls{ml} classifiers.
%Printed bespoke architectures leverage the low \gls{nre} costs and high degree of customization, making it feasible to implement on-demand, resource-efficient \gls{ml} classifiers.
Additionally, exploiting the intrinsic error resilience of ML circuits~\cite{Henkel:ICCAD2022:expedition}, state-of-the-art approaches combine bespoke circuits with approximate computing to boost hardware efficiency~\cite{Armeniakos:DATE2022:axml,Armeniakos:TCAD2023:cross,Armeniakos:TC2023:codesign,Afentaki:ICCAD23:hollistic,Afentaki:DATE2024:embedding,Mrazek:ICCAD2024}.
However, they often lead to significant accuracy loss in order to meet the tight power and area constraints inherent in PE. 

% % Printed SVMs - Limitations of SOTA
% A wide range of printed \gls{ml} classifiers have been showcased in the literature, ranging from \glspl{dt}~\cite{Balaskas:ISQED2022:axDT} and \glspl{svm}~\cite{Mubarik:MICRO:2020:printedml, Armeniakos:DATE2022:axml, Armeniakos:TCAD2023:cross} to \glspl{mlp}~\cite{Afentaki:DATE2024:embedding, Afentaki:ICCAD23:hollistic,Kokkinis:DATE2023:minimization}.
% At the higher end of the area consumption spectrum stand more complex models (e.g., \glspl{mlp}), which may be sub-optimal in delivering ultra-low area for printed circuits.
% Printed \glspl{svm} on the other hand offer much lower area at comparable accuracy levels in the form of regression-based~\cite{Mubarik:MICRO:2020:printedml}, classification-based~\cite{Armeniakos:TCAD2023:cross} and \gls{lut}-based architectures~\cite{Mubarik:MICRO:2020:printedml}.
% All the above constitute fully-parallel classifiers, opting for memory-less implementations, however without taking hardware re-use into account as an area reduction mechanism.

% Our solution: sequential SVMs
A wide range of digital printed \gls{ml} classifiers have been showcased in the literature, with the main focus lying on fully parallel implementations of shallow Neural Networks~\cite{Armeniakos:TC2023:codesign,Afentaki:ICCAD23:hollistic,Afentaki:DATE2024:embedding,Mrazek:ICCAD2024}. 
In this work, we leverage the high accuracy of \glspl{svm} in target applications and propose, for the first time, \emph{printed sequential \gls{svm} classifiers}.
Our goal is to minimize area while maintaining high accuracy.
We design a compute engine with just one \gls{mac}, folding each support vector computation over it, maximizing hardware reuse and significantly reducing area.
Additionally, we reduce the area requirements associated with sequential elements by minimizing the use of registers--which are costly in PE--and we use bespoke multiplexers (MUXs) for storing model parameters.
A bespoke control unit further folds the entire \gls{svm} computation over the single support vector engine.
We implement the One-vs-One (OvO) algorithm~\cite{ovo} into a binary decision tree, simplifying the control circuitry and eliminating the need for a voter and additional registers.
Compared to printed parallel SVMs, our sequential ones achieve more than $5$x lower area at similar accuracy.
\textbf{Our novel contributions within this work are as follows}:
\begin{enumerate}[topsep=0pt,leftmargin=*]
\item We propose, design, and evaluate, for the first time, printed sequential \gls{svm} circuits targeting printed \gls{ml} classification.
The hardware description of our \glspl{svm} is automatically generated\footnote{Our circuits are available on:
\href{https://github.com/floAfentaki/Support-Vector-Machine-Circuits-Targeting-Printed-Electronics/tree/main/Compact-And-Highly-Accurate-Printed-Sequential-SVMs_ISCAS2025}{https://github.com/floAfentaki/Support-Vector-Machine-Circuits-Targeting-Printed-Electronics}}. 
\item
% To the best of our knowledge, 
\blue{Our \glspl{svm} constitute} the most accurate digital printed \gls{ml} classifiers that feature acceptable area and power overheads, and also meet the strict physical constraints and limited power sources of PE. 
\end{enumerate}

\section{Related Work}
\label{sec:elated}
In recent years, research on printed \gls{ml} classifiers has proliferated, with a primary focus on improving hardware efficiency while adhering to the strict area and power budget of \gls{pe} systems.
As a result, classification accuracy is often compromised in favor of design feasibility.
The authors in~\cite{Mubarik:MICRO:2020:printedml} introduced printed \gls{ml} classifiers by exploring various models and architectures, and concluding that only simpler models like decision trees and \gls{svm} regressors could be implemented in \gls{pe}.
Furthermore,~\cite{Mubarik:MICRO:2020:printedml} determined that fully parallel architectures should be used in \gls{pe}. %, while sequential designs can be prohibitive due to the costs associated with registers.
Since then, significant research efforts have focused on fully parallel approximate neural networks (\glspl{mlp}~\cite{Armeniakos:DATE2022:axml, Armeniakos:TCAD2023:cross, Armeniakos:TC2023:codesign,Afentaki:ICCAD23:hollistic,Afentaki:DATE2024:embedding} and \glspl{bnn}~\cite{Mrazek:ICCAD2024}) and also approximate \glspl{svm} classifiers~\cite{Armeniakos:DATE2022:axml, Armeniakos:TCAD2023:cross}.
%Approximate computing has proven effective in reducing complexity, with approaches spanning hardware, software, and co-design techniques.
While works like~\cite{Afentaki:ICCAD23:hollistic,Afentaki:DATE2024:embedding,Mrazek:ICCAD2024} achieve impressive area efficiency, they often report accuracy losses around 4\%-5\% or more, due to employing aggressive power-of-two or ternary quantization and approximate additions.
On the other hand,~\cite{Armeniakos:DATE2022:axml} introduced a more conservative post-training approximation technique involving hardware-friendly weight replacement to approximate the multiplications in \glspl{svm} and \glspl{mlp}.
\cite{Armeniakos:DATE2022:axml} also applied a gate-level pruning approximation.
The authors in~\cite{Armeniakos:TCAD2023:cross} expand upon~\cite{Armeniakos:DATE2022:axml} by additionally incorporating voltage over-scaling.
In-training hardware-friendly weight replacement along with truncated addition is proposed in~\cite{Armeniakos:TC2023:codesign} for \glspl{mlp}.
%Due to the more conservative approximation nature in~\cite{Armeniakos:DATE2022:axml,Armeniakos:TCAD2023:cross,Armeniakos:TC2023:codesign} acceptable accuracy is mostly achieved. 
Similar to the aforementioned works, we consider the Electrolyte-Gated FET (EGFET) technology~\cite{Bleier:ISCA:2020:printedml}, which offers good mobility characteristics and operation at low supply voltages ($0.6$V-$1$V~\cite{Marques:Materials:2019}), aligning well with printed battery-powered applications. 

% Our work differs from existing approaches by designing the first printed sequential \gls{svm} classifiers, demonstrating for the first time that digital printed \gls{ml} classifiers can meet the power and area constraints of \gls{pe} with no loss in accuracy.
Our work distinguishes from existing approaches by introducing the first design of printed sequential \gls{svm} classifiers, demonstrating for the first time that digital printed classifiers can meet the physical constraints of \gls{pe} with high accuracy.

\section{Printed Bespoke Sequential SVMs}
\label{sec:svm}
\begin{figure}[!t]
    \centering
    \includegraphics[width=\columnwidth]{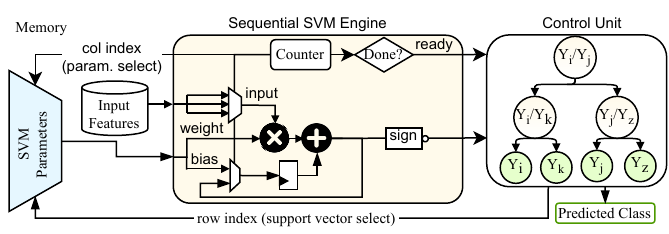} \vspace{-4ex}
    \caption{\blue{Overview of our proposed sequential \gls{svm} circuits.
    They consist of three main components memory, SVM compute engine and control unit.}}
    \label{fig:architecture}\vspace{-3ex}
\end{figure}

\gls{svm} is a supervised learning algorithm able to classify data by determining the optimal hyperplane for separating classes in a high-dimensional feature space.
\glspl{svm} are effective in handling small, high-dimensional datasets, offering robustness against overfitting.
In brief, \glspl{svm} compute a number of support vectors
%, one for each pair of classes,
and based on the obtained results, determine the class with the most wins.
In printed hardware, linear kernels are typically preferred within support vectors, to maximize area efficiency.
Existing implementations~\cite{Mubarik:MICRO:2020:printedml, Armeniakos:TCAD2023:cross} design fully parallel bespoke circuits, where model parameters are hardwired into the circuit implementation, eliminating the need for costly sequential elements in \gls{pe}, but requiring a hardware multiplier for each \gls{svm} weight.
In contrast, our work focuses on designing sequential \glspl{svm} to minimize the required arithmetic units, making it crucial, nevertheless, to optimize the cost of memory and registers.
Notably, in CMOS technologies, a D Flip-flop's area-equivalent is $4$ NAND gates, whereas in EGFET, this number rises to $6$--a $50$\% increase.

An abstract overview of our proposed sequential SVM circuit is shown in~\cref{fig:architecture}.
It consists of three main components, i.e., parameter storage, control, and compute, all working in sync to fold the entire \gls{svm} prediction over a single \gls{mac} unit. 
Each iteration involves fetching the respective support vector from memory (as determined by the control unit) and computing it in the support vector engine (a multi-cycle operation over a single \gls{mac}).
The result is fed back to the control unit, which either advances the computation (intermediate step) or selects the winning class (final step).
Importantly, our design choices for implementing our sequential \gls{svm} circuits target area efficiency.
Nevertheless, since power consumption in EGFET circuits is mostly static and internal (short-circuit), minimizing area also effectively reduces power.

\subsection{SVM Training} \label{sec:train}
Two established strategies for SVM multi-class classification are OvO and OvA (One-vs-All)~\cite{bishop2006pattern:ovo_ova}.
In OvO, each binary classifier distinguishes between two classes, requiring $\frac{n(n-1)}{2}$ classifiers for $n$ classes.
OvA uses $n$ classifiers, each separating one class from the rest.
OvO requires storing more support vectors but works with smaller subsets of training data, avoiding accuracy loss from imbalanced datasets.
Prioritizing high accuracy and relying on the inherent area efficiency of our sequential implementation, we choose OvO.
For the examined datasets (Section~\ref{sec:eval}), using quantized support vectors, OvO delivers $8.7$\% higher accuracy, on average, compared to OvA.

%Training is performed using scikit-learn's LinearSVC class, with randomized parameter optimization, to train a classifier for each possible class pair.
%Inputs are normalized to $[0,1]$, and training/testing is done with a random $80$\%/$20$\% split.
Training is performed using scikit-learn's LinearSVC class to train a classifier (support vector) for each possible class pair.
Inputs are truncated to $4$-bit fixed-point, and post-training, weights and biases are quantized with min-max linear scaling to the lowest precision that results in negligible accuracy loss (within $0.5$\%).
Additionally, SVM inference is profiled to determine the minimum precision needed for the support vector engine’s accumulator.
The Verilog description for each trained SVM is automatically generated using code templates.
The extracted precisions and SVM parameters are stored in a configuration file.
The support vector engine is the same, with only precision adjustments, while a unique control unit is generated for each model.

\subsection{Support Vector Storage}\label{sec:memory}
Sequential architectures must store model parameters in memory (weights and biases in our \glspl{svm}).
A printed \gls{rom} is a prominent choice for this.
A compact \gls{rom} was proposed and fabricated in~\cite{Bleier:ISCA:2020:printedml}, outperforming other state-of-the-art designs.
It uses a crossbar architecture, where cross-points are shorted by printing conductive material (such as PEDOT:PSS) to represent bit values.
By varying the geometry of the conductive material, multi-bit values can be stored in a single printed dot.
An analog-to-digital converter (ADC) is needed to read stored values.
Considering the cost of ADCs and ROM cells in~\cite{Bleier:ISCA:2020:printedml}, we deduce that storing \mbox{$2$-bit} values is optimal for reading $2$-bit to $8$-bit words as for our model parameters.
This assumes the use of one to four \mbox{$2$-bit} ADCs to maintain reasonable performance.
Since our design uses a single-MAC \gls{svm} engine, we need to process only a single model parameter (weight or bias) per cycle.
One support vector is stored per crossbar row.
Based on the selected support vector (from the control unit), the respective row is activated. 
Then, a set of columns, depending on the precision of the model parameters, are activated each cycle.
A counter (see Section~\ref{sec:engine}) that selects the appropriate parameter is used as the column index.
Nevertheless, such an architecture is highly vulnerable to printing variations, which can alter stored parameters and lead to significant accuracy degradation~\cite{Zhao:DATE2023}. 

As a more robust alternative, we propose the use of bespoke MUXs.
The MUX input data are hardwired to the model parameters, with the row and column indexes described above serving as the input select signals.
While this approach offers improved robustness, it is less dense than ROM storage.
For example, for the Pendigits dataset (see Section\ref{sec:eval}) that features $45$ support vectors with $18$ $8$-bit parameters each, the area, power, and latency of the ROM are $2.4$cm$^2$, $1.9$mW, and $15$ms, respectively.
The corresponding values for MUX-based storage are $5.0$cm$^2$, $5.6$mW, and $19$ms, respectively.
Although less hardware-efficient, the overhead of the MUX-based storage is not prohibitive, especially considering that this example represents the worst-case difference among all datasets examined.
Therefore, we choose bespoke MUX-based storage for our SVMs to prioritize accuracy and robustness.

% \noindent\textbf{MAC-Based Computational Engine:}
\subsection{Single MAC Support Vector Engine} \label{sec:engine}
Our control unit makes decisions by evaluating one support vector at a time.
Hence, we design a support vector engine to compute the corresponding output:
\begin{equation}
    y = \begin{cases}
        1 & \text{if}\, \sum_{i=1}^{m} w_i x_i + b \geq 0 \\
        0 & \text{otherwise,}
    \end{cases}
\label{eq:compute_mac}
\end{equation}
where $w_i$ and $b$ are the support vector's weights and bias, and $x_i$ are the input activations.
Striving for area efficiency, we fold the computation of \eqref{eq:compute_mac} over a single MAC unit.

As shown in \cref{fig:architecture}, a register stores the accumulation result, whose size is minimized based on partial sum profiling during SVM training.
A second small register is used to store a counter, which serves as the column index for the memory and for synchronization.
In the first cycle, the bias is fetched, initializing the accumulation register.
In each subsequent cycle, a weight is read from memory, multiplied by the respective input, and the product is accumulated.
Once all model parameters are processed, the engine outputs a ready signal, providing the inverted sign of the accumulation.

\begin{figure}[!t]
    \centering
    \includegraphics[width=.9\columnwidth]{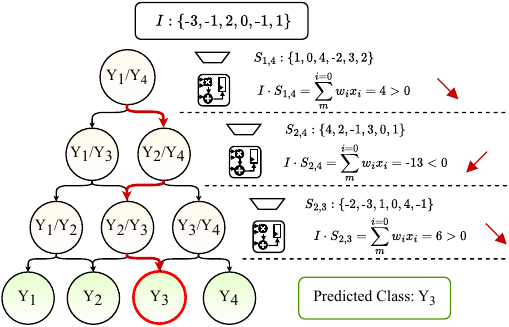}
    \caption{Example execution of our sequential \glspl{svm}. $4$ classes and $6$ input features/weights are considered. Support vectors corresponding to selected nodes are shown on the right, along with the classification output of the support vector engine. The output class is predicted in $3\times 7=21$ cycles. Fixed point values are represented by integers.
    }
    \label{fig:example}\vspace{-4ex}
\end{figure}

% \noindent\textbf{Control Unit:}
\subsection{Control Unit} \label{sec:control_unit}
OvO computes a support vector for each pair of classes, with the class attaining the highest score being selected as the SVM output.
In~\cite{Armeniakos:DATE2022:axml,Armeniakos:TCAD2023:cross}, all support vectors are computed in parallel, and a voter circuit determines the winning class.
While this approach requires a voter, a sequential implementation also requires a log$_2\frac{n(n-1)}{2}$ bit counter and additional registers to track the winning class and its score.
To minimize register use, we implement OvO as a decision-directed acyclic graph (DDAG)~\cite{ovo}--specifically, a binary decision tree--for our control \gls{fsm}, requiring only a small register of log$_2\frac{n(n-1)}{2}$ bits to store the FSM state.
%A software implementation of OvO as a DDAG is detailed in~\cite{ovo}, where each state represents the current winning class.
Each DDAG node (FSM state) represents the current winning class, and is linked to a support vector that compares this class against a new class, yet to be considered.
FSM selects the corresponding support vector by its row index in the memory.
%According to its row index, the support vector is fetched from memory.
After $m+1$ cycles (where $m$ is the number of input features), the support vector is computed in our engine, and the winner of this comparison is determined.
The newly identified winning class becomes the current winner, prompting the DDAG to transition to the right if the new class prevails, or to the left otherwise.

The hardware implementation of the FSM is straightforward.
It uses a hardcoded row index per state and only requires a MUX to select between two predefined next states based on the support vector engine's output.
\cref{fig:example} presents a detailed example of our sequential execution and FSM flow.
The FSM has $\frac{n(n-1)}{2}$ states (the number of OvO support vectors) but requires only $n-1$ support vector evaluations, resulting in a total of $(n-1) \times (m+1)$ cycles for each classification.

\section{Results and Analysis} \label{sec:eval}

In this section, we evaluate our printed sequential SVM classifiers in terms of accuracy and hardware overhead.
The evaluation is based on five datasets (see Table~\ref{tab:svmfp}) from the UCI ML repository~\cite{Dua:2019:uci_datasets}, selected because they involve sensor inputs relevant to printed applications~\cite{Mubarik:MICRO:2020:printedml} and are frequently used by the state of the art.
We use Synopsys Design Compiler and PrimeTime for hardware evaluation, targeting frequency in the Hz range, typical for printed applications~\cite{Bleier:ISCA:2020:printedml}.
The fully parallel SVM classifiers--designed as described in~\cite{Mubarik:MICRO:2020:printedml}--serve as our baselines.
Additionally, we compare against~\cite{Armeniakos:TCAD2023:cross,Armeniakos:TC2023:codesign}, which, among all existing approximate printed digital \gls{ml} classifiers, achieve relatively high accuracy across all datasets.
To train our SVMs we use randomized parameter optimization, inputs are normalized to $[0,1]$, and training/testing is done with a \blue{randomly distributed $80$\%/$20$\% split}.
All accuracies hereafter are reported on the test dataset.
\blue{Table~\ref{tab:svmfp} presents the details of our trained SVMs}, reporting also the FP32 software accuracy.

Table~\ref{tab:compsoa} presents the hardware evaluation of our sequential SVMs \blue{in comparison to state-of-the-art designs}.
As shown, \blue{our SVMs exhibit negligible to no accuracy loss compared to the FP32 model.}
The average area of our designs is $3.7$cm$^2$, with a maximum of $7.8$cm$^2$, making them realistic and suitable for most printed applications~\cite{Armeniakos:TCAD2023:cross}. 
Similarly, the average power consumption is $4.2$mW, with a maximum of $8.7$mW, enabling all our circuits to be powered by existing printed batteries, e.g., Zinergy $15$mW or BlueSpark $6$mW.
% Table~\ref{tab:compsoa} demonstrates the feasibility and practicality of our SVMs while maintaining high accuracy.
Table~\ref{tab:compsoa} demonstrates the feasibility and practicality of our highly accurate SVMs.
Even when considering the power required for ADCs to process sensor data (e.g., $0.33$mW for the required $4$-bit ADC~\cite{Bleier:ISCA:2020:printedml}), our SVMs can still operate within printed battery constraints.
Since our circuits process one input feature per cycle, a single ADC can be shared across many input sensors.
In contrast, ADC costs might become bottlenecks in parallel designs like~\cite{Mubarik:MICRO:2020:printedml, Armeniakos:TCAD2023:cross, Armeniakos:TC2023:codesign}, \blue{where inputs are processed simultaneously}.
% which require all inputs simultaneously, might face bottlenecks due to ADC costs in their parallel designs. 
We also evaluated our SVMs \blue{using a printed crossbar ROM for storage}, which would reduce the area for Cardio, Dermatology, Pendigits, RedWine, and WhiteWine by $10$\%, $23$\%, $28$\%, $8$\%, and $3$\%, respectively.
However, to fully exploit these (limited) gains, a variation-aware training approach, e.g., similar to~\cite{Zhao:DATE2023}, would be necessary, which does not consistently yield reliable results across all datasets.

\begin{table}[t!]
\setlength\tabcolsep{3pt} 
\caption{SVM model analysis for the examined datasets.}
\label{tab:svmfp}
\footnotesize
\centering
\renewcommand{\arraystretch}{1}
%\begin{adjustbox}{width=\textwidth} % Automatically resize table to fit page width

\begin{tabular}{l| S[table-format=2.1]S[table-format=2.0]S[table-format=2.0]S[table-format=2.0]}

\hline
\textbf{Dataset} & \textbf{FP32 Acc (\%)} & \textbf{\#Inputs} & \textbf{\#Classes} & \textbf{\#Support Vectors}\\ \hline
Cardio & 93.4 & 21 & 3  & 3 \\
Dermatology & 98.6 & 33 & 6  & 15 \\ 
Pendigits & 97.7 & 17 & 10 & 45 \\
RedWine   & 66.2 & 11 & 6  & 15 \\
WhiteWine & 56.6 & 11  & 7 & 21 \\ \hline
\end{tabular}\vspace{-5ex}

\end{table}

Compared to fully parallel exact SVMs~\cite{Mubarik:MICRO:2020:printedml}, as shown in Table~\ref{tab:compsoa}, our sequential SVMs achieve $10$x lower area and $30$x lower power, on average.
For some datasets, our SVMs achieve higher accuracy due to our use of the more flexible LinearSVC class (compared to the SVC used in~\cite{Mubarik:MICRO:2020:printedml}), which provides better scalability and a wider range of penalty and loss functions.
Against the approximate SVMs~\cite{Armeniakos:TCAD2023:cross}, our circuits exhibit, on average, $6$x and $12$x lower area and power, respectively, while attaining $4.8$\% higher accuracy.
Compared to approximate MLPs~\cite{Armeniakos:TC2023:codesign}, our SVMs yield $2$x less area and $6$x less power on average, with $6$\% higher accuracy.
For RedWine, \cite{Armeniakos:TC2023:codesign} demonstrates a lower area, but both circuits are small-enough, whereas our design achieves $10\%$ higher accuracy.
Compared to more approximate digital printed neural networks~\cite{Mrazek:ICCAD2024},~\cite{Afentaki:ICCAD23:hollistic} (not included in Table~\ref{tab:compsoa}), our SVMs feature higher area but achieve, on average, $7.4$\% and $7.9$\% higher accuracy than~\cite{Mrazek:ICCAD2024} and~\cite{Afentaki:ICCAD23:hollistic}, respectively.
Interestingly, for the Pendigits dataset (the most complex examined), our SVMs offer $6$x lower area, $5$x lower power, and $4$\% higher accuracy than~\cite{Mrazek:ICCAD2024}, and $3$x lower area and power, and $8$\% higher accuracy compared to~\cite{Afentaki:ICCAD23:hollistic}.
Overall, as dataset complexity increases, the advantages of our sequential approach become even more pronounced, with higher gains over the state-of-the-art.
% This highlights the critical impact of our approach, making it necessary for achieving high accuracy in more complex datasets.
Furthermore, it's important to note that, except for RedWine and WhiteWine~\cite{Armeniakos:TC2023:codesign}, all state-of-the-art classifiers in Table~\ref{tab:compsoa} consume over $30$mW, which does not align with printed batteries~\cite{Mubarik:MICRO:2020:printedml}, and would be impractical for printed applications.
Concluding, within the strict physical constraints of printed applications in area and power availability, our SVMs constitute the most accurate printed classifiers that successfully meet these requirements.

\begin{table}[t!]
\setlength\tabcolsep{4pt} 
\caption{Hardware evaluation and comparison with state of the art.}
\label{tab:compsoa}
\footnotesize
\centering
\renewcommand{\arraystretch}{1}
%\begin{adjustbox}{width=\textwidth} % Automatically resize table to fit page width

\begin{tabular}{ll|S[table-format=2.1]|S[table-format=3.1]S[table-format=3.1]S[table-format=2]}

\hline
\textbf{Dataset} & \textbf{Technique} & {\thead{\textbf{Accuracy}\\ (\%)}} & {\thead{\textbf{Area} \\ (cm$^{2}$)}} & {\thead{\textbf{Power}\\ (mW)}} & {\thead{\textbf{Freq.}\\ (Hz)}}\\ \hline
Cardio & Parl SVM~\cite{Mubarik:MICRO:2020:printedml} & 90.0 & 15.1 & 57.4 & 13 \\
Cardio & Ax Parl SVM~\cite{Armeniakos:TCAD2023:cross} & 89.0 & 17.0 & 48.9 & 13 \\
Cardio & Ax Parl MLP~\cite{Armeniakos:TC2023:codesign} & 87.0 & 6.1 & 20.8 & 5 \\
Cardio & \oursTable{Ours Seq. SVM} & 93.1 & 3.1 & 3.6 & 24 \\
\hline
Dermatology & Parl SVM~\cite{Mubarik:MICRO:2020:printedml} & 97.2 & 60.4 & 182.9 & 8 \\
Dermatology & \oursTable{Ours Seq. SVM} & 98.6 & 4.9 & 5.3 & 18 \\
\hline
Pendigits & Parl SVM~\cite{Mubarik:MICRO:2020:printedml} & 97.8 & 123.8 & 364.4 & 4 \\
Pendigits & Ax Parl SVM~\cite{Armeniakos:TCAD2023:cross} & 97.0 & 97.0 & 183.7 & 4 \\
Pendigits & Ax Parl MLP~\cite{Armeniakos:TC2023:codesign} & 93.0 & 32.7 & 99.2 & 4 \\
Pendigits & \oursTable{Ours Seq. SVM} & 97.6 & 7.8 & 8.7 & 15 \\
\hline
RedWine & Parl SVM~\cite{Mubarik:MICRO:2020:printedml} & 57.0 & 23.5 & 92.8 & 15 \\
RedWine & Ax Parl SVM~\cite{Armeniakos:TCAD2023:cross} & 56.0 & 11.7 & 21.3 & 15 \\
RedWine & Ax Parl MLP~\cite{Armeniakos:TC2023:codesign} & 56.0 & 1.1 & 3.9 & 5 \\
RedWine & \oursTable{Ours Seq. SVM} & 66.2 & 3.1 & 3.5 & 19 \\
\hline
WhiteWine & Parl SVM~\cite{Mubarik:MICRO:2020:printedml} & 53.0 & 28.3 & 112.4 & 17 \\
WhiteWine & Ax Parl SVM~\cite{Armeniakos:TCAD2023:cross} & 52.0 & 11.0 & 34.7 & 17 \\
WhiteWine & Ax Parl MLP~\cite{Armeniakos:TC2023:codesign} & 53.0 & 6.5 & 21.3 & 5 \\
WhiteWine & \oursTable{Ours Seq. SVM} & 56.2 & 3.4 & 3.9 & 20 \\
\hline
\end{tabular}\vspace{-4ex}

\end{table}

As shown in Table~\ref{tab:compsoa}, folding the SVM execution allows for higher clock frequencies.
However, this frequency gain does not fully offset the increased latency and energy demands due to the additional cycles required.
Nevertheless, this is unlikely to be an issue for most applications, as performance is not the primary focus in low-throughput printed applications~\cite{Henkel:ICCAD2022:expedition}.
Similarly, printed batteries are customizable in capacity, shape, and voltage~\cite{PrintedBatteries2018}, so energy concerns might be secondary compared to power availability.
\section{Conclusion} \label{sec:conclusion}

In this paper, we present the first printed sequential SVM classifiers optimized for the strict area and power constraints of printed electronics (PE).
Using our proposed sequential architecture with bespoke storage and control, and a single-MAC engine, we achieve on average $10$x lower area and $30$x lower power compared to fully-parallel exact SVMs, at similar accuracy.
Compared to approximate printed classifiers, our approach reduces area and power by $4$x and $9$x, respectively, on average, while delivering a $5.4$\% accuracy gain.
These advantages become more pronounced as dataset complexity increases.
Overall, our SVM circuits represent the most accurate digital printed classifiers that meet PE's strict area and power constraints.
Although we focused on EGFET technology, our design principles are applicable to similar technologies.

\blue{\section*{Acknowledgment}
This work is funded by the H.F.R.I call “Basic research Financing (Horizontal support of all Sciences)” under the National Recovery and Resilience Plan “Greece 2.0” (H.F.R.I. Project Number: 17048).}


\end{document}